\documentclass[letterpaper]{article}
\usepackage{aaai25} 
\usepackage{times} 
\usepackage{helvet} 
\usepackage{courier} 
\usepackage[hyphens]{url} 
\usepackage{graphicx} 
\usepackage{graphicx} 
\usepackage{natbib} 
\usepackage{caption} 
\usepackage{amsmath}
\usepackage{footnote}
\usepackage{xcolor}
\usepackage{booktabs}
\usepackage{multirow}
\usepackage{multicol}
\usepackage{colortbl }
\usepackage{amssymb}
\usepackage{algorithm}
\usepackage{algorithmic}
\title{Fast Omni-Directional Image Super-Resolution: Adapting the Implicit Image Function with Pixel and Semantic-Wise Spherical Geometric Priors}
\author {
    Xuelin Shen\textsuperscript{\rm 2}\footnote{These authors contributed equally.},
    Yitong Wang\textsuperscript{\rm 1,2}\footnotemark[1],
    Silin Zheng\textsuperscript{\rm 1},
    Kang Xiao\textsuperscript{\rm 2},
    Wenhan Yang\textsuperscript{\rm 3},
    Xu Wang\textsuperscript{\rm 1}\footnote{Corresponding author.}
}
\affiliations {
    \textsuperscript{\rm 1}College of Computer Science and Software Engineering, Shenzhen University\\
    \textsuperscript{\rm 2}Guangdong Laboratory of Artificial Intelligence and Digital Economy(SZ)\\
    \textsuperscript{\rm 3}Peng Cheng Laboratory\\
    shenxuelin@gml.ac.cn,
    ted.yitongwang@gmail.com,
    zhengpny@gmail.com,
    xiaokang2022@email.szu.edu.cn,
   yangwh@pcl.ac.cn,
    wangxu@szu.edu.cn
}

\begin{document}

\maketitle

\begin{abstract}
In the context of Omni-Directional Image (ODI) Super-Resolution (SR), the unique challenge arises from the non-uniform oversampling characteristics caused by EquiRectangular Projection (ERP).
Considerable efforts in designing complex spherical convolutions or polyhedron reprojection offer significant performance improvements but at the expense of cumbersome processing procedures and slower inference speeds.
Under these circumstances, this paper proposes a new ODI-SR model characterized by its capacity to perform Fast and Arbitrary-scale ODI-SR processes, denoted as FAOR.
The key innovation lies in adapting the implicit image function from the planar image domain to the ERP image domain by incorporating spherical geometric priors at both the latent representation and image reconstruction stages, in a low-overhead manner.
Specifically, at the latent representation stage, we adopt a pair of pixel-wise and semantic-wise sphere-to-planar distortion maps to perform affine transformations on the latent representation, thereby incorporating it with spherical properties.
Moreover, during the image reconstruction stage, we introduce a geodesic-based resampling strategy, aligning the implicit image function with spherical geometrics without introducing additional parameters.
%
%
As a result, the proposed FAOR outperforms the state-of-the-art ODI-SR models with a much faster inference speed.  
Extensive experimental results and ablation studies have demonstrated the effectiveness of our design.
\end{abstract}

\begin{links}
    \link{Code}{https://github.com/GingaUL/FAOR}
\end{links}

\begin{figure}[!ht]
    \centering
    \includegraphics[width=0.4\textwidth]{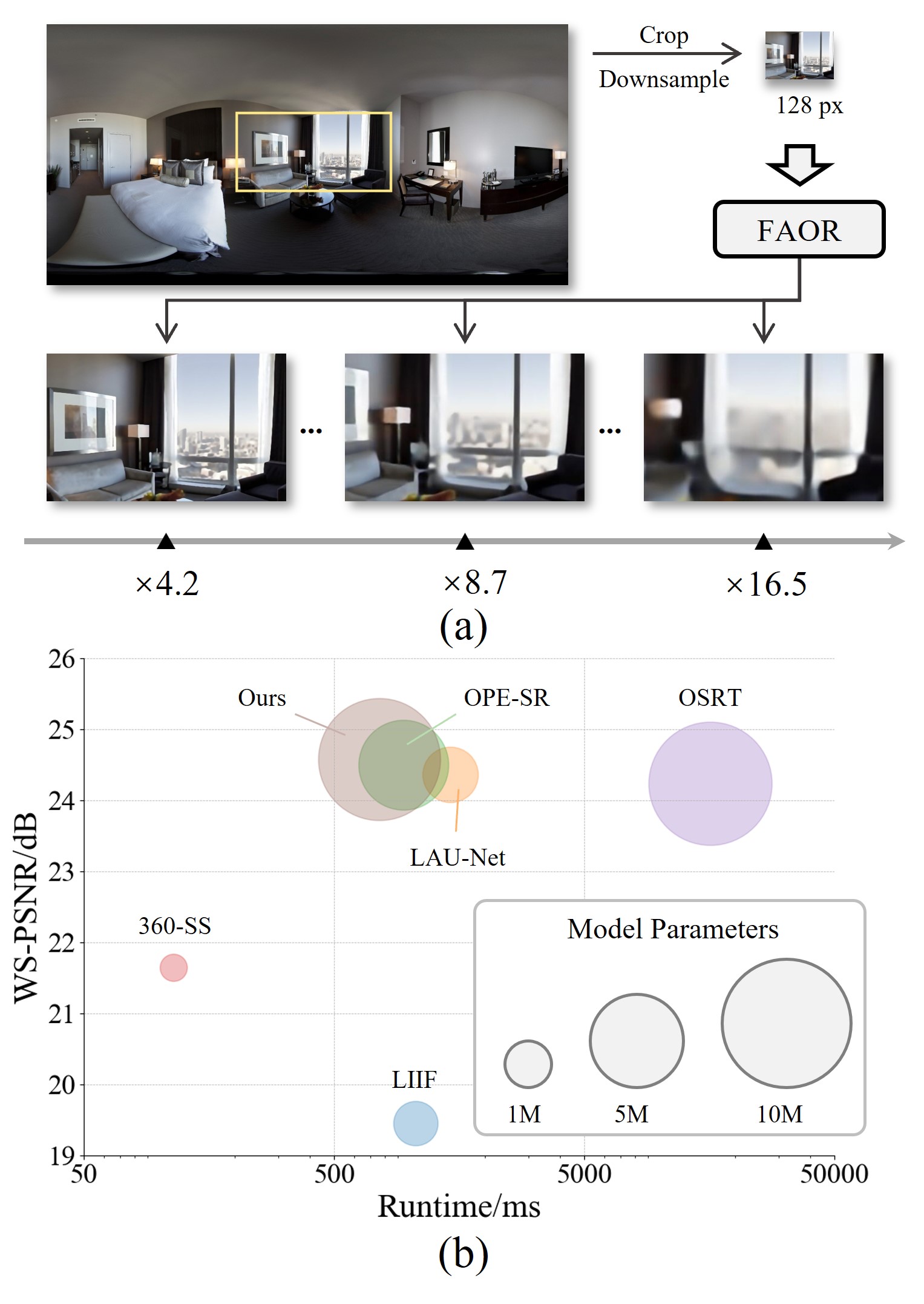}
    \caption{(a) The proposed FAOR provides a continuous ODI representation, enabling arbitrary-scale SR. (b) Comparisons with other state-of-the-art models regarding performance and inference speed on the $\times 8$ SR task}
    \label{fig:Arbitrary-Performance}
\end{figure}

\section{Introduction}
Characterized by a wide $360^{\circ} \times 180^{\circ}$ field of view, Omni-Directional Images (ODIs) possess unique advantages in providing more comprehensive scene representations and immersive viewing experiences compared to conventional planar images.
As a result, recent days have witnessed the blooming applications of ODIs in both machine and human vision-oriented domains, \textit{e.g.}, augmented reality (AR), virtual reality (VR), autonomous driving~\cite{yang2021capturing, PANDORA}, and robot navigation~\cite{yang2020graph,kang2019interactive}.
In practice, ODIs are usually stored and transmitted in Low Resolution (LR) to satisfy the requirements of real-time machine vision-oriented applications, and subsequently undergo Super-Resolution (SR) to achieve high perceptual quality for human vision.

As for ODI-SR, the main challenges arise from the distortion caused by sphere-to-planar projection. 
Specifically, raw ODIs are typically stored in EquiRectangular Projection (ERP) format, which straightforwardly maps longitude to equally spaced columns and latitude to equally spaced rows.
As a result, the pixels in high-latitude regions are oversampled, leading to significant content deformation in these areas.
As such, directly adapting planar image SR models to ODI images is inappropriate, as they cannot comprehensively capture the scene's contextual information without considering ODI-specific characteristics.
In addressing this challenge, numerous efforts have been devoted to various strategies, \textit{e.g.,} developing specific convolution operators aligned with spherical characteristics~\cite{lee2019spherephd}, leveraging latitude-adaptive methodologies that allow different latitude regions to adopt distinct upscaling factors~\cite{deng2021lau,cai2024spherical}, or introducing new polyhedron-based reprojection methods to minimize sphere-to-planar distortion~\cite{yoon2022spheresr}.
Although significant achievements have been made, these ODI-oriented modules involve cumbersome implementation procedures, lead to time-consuming inference, and present significant obstacles to practical implementation.
This motivates us to explore a more efficient approach that incorporates spherical characteristics into the SR process, pursuing both performance and running speed.

To this end, this paper proposes a new ODI-SR method that adapts the implicit image function to the ERP domain by incorporating the sphere geometric priors to both the \textit{latent representation} and \textit{image reconstruction} stages.
In general, the implicit image function aims at establishing map functions between the coordinates and corresponding latent representations to pixel values, favored by its capacity for continuous image representation and fast inference speed~\cite{chen2021learning,song2023ope}.
Herein, at the \textit{latent representation} stage, we first explore leveraging affine transformation for sphere-to-planar distortion representation, enjoying the streamlined computation with standard image operators. 
Thus, we propose the Affine-Transformation-based Feature Modulating (ATFM) module and insert it into the feature encoder, within which the affine transformations are applied to the extracted ERP features, obtaining sphere-to-planar aware latent representations.
Moreover, the ATFM module is guided by a set of external priors, including a pixel-wise stretching ratio map and an instance segmentation map, which are responsible for providing insights into the sphere-to-planar distortions from pixel-wise and semantic perspectives, respectively.
%
%
%
%
During the \textit{image reconstruction} stage, we introduce a spherical geodesic-based resampling function that leverages unbiased spherical locations for the latent representation resampling process.
We highlight our main contributions as follows: 
\begin{itemize}
    \item To the best of our knowledge, we are the first to address ODI-SR by jointly considering performance and running speed. With architectures specially designed to capture spherical characteristics and ensure inference simplicity, the proposed FAOR is capable of outperforming existing methods while achieving significantly faster running speeds. as illustrated in Fig.~\ref{fig:Arbitrary-Performance}.
    \item We explore the affine transformation for representing sphere-to-planar distortion, incorporating pixel-wise and semantic-wise spherical geometric priors into the ODI latent representation.
    \item A Spherical Geodesic-based Implicit Image Function (SGIF) is proposed to provide a continuous and spherically unbiased ODI representation, facilitating efficient and arbitrary-scale ODI-SR processes.
\end{itemize}

\begin{figure*}[ht]
    \centering    \includegraphics[width=0.92\textwidth]{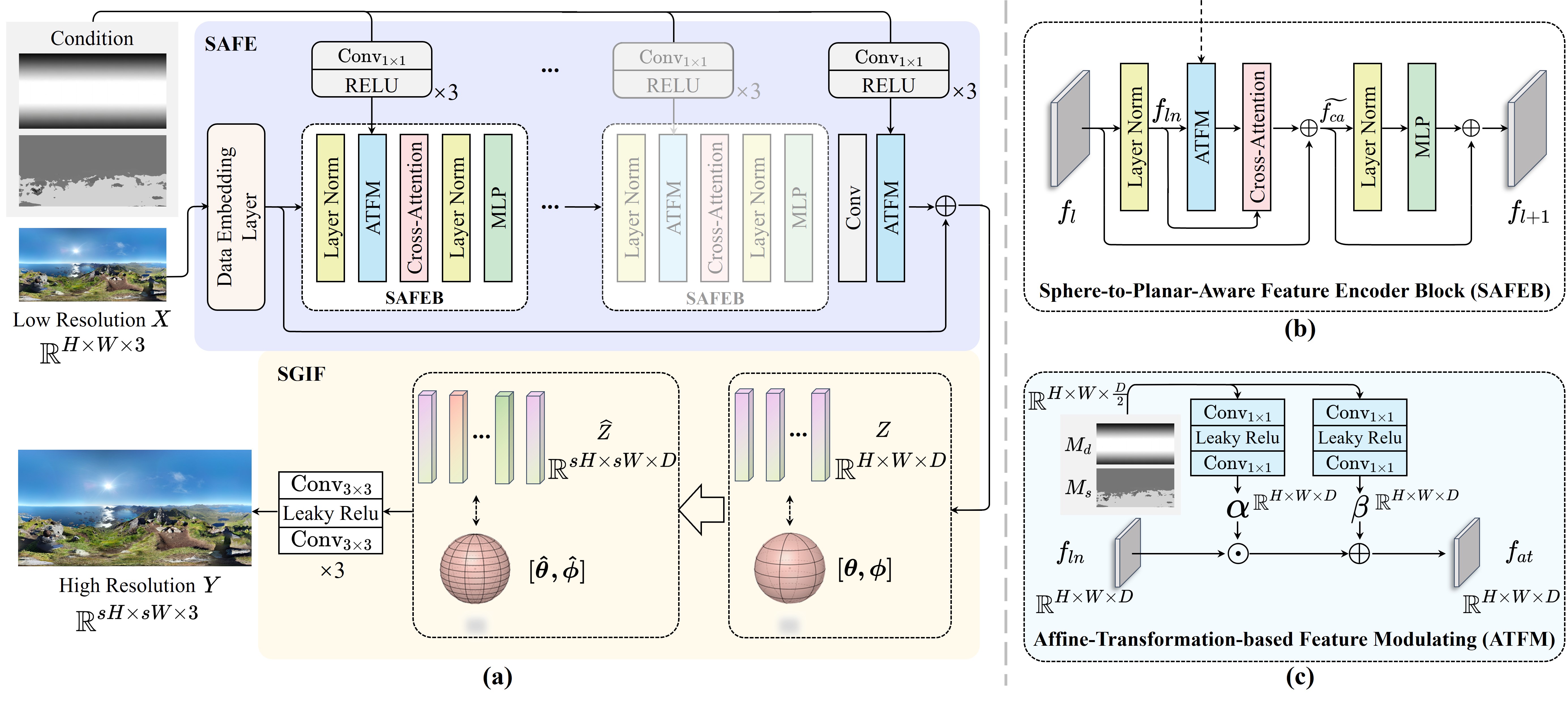}
    \caption{Overall architectures of the proposed FAOR and detailed design of key modules.}
    \label{fig:framework}
\end{figure*}

\section{Related Works}
\label{sec_background}
\subsection{Omni-Directional Image Machine Vision}
ODI-oriented machine vision research emphasizes understanding the object deformation caused by non-uniform sampling due to sphere-to-planar projection.
Preliminary works focused on modifying and adapting regular convolution kernels to produce deformation-aware features.
In special, Su \textit{et al.}~\cite{su2017learning} made the first attempt that leverages the knowledge distillation to obtain adaptive kernel sizes for 2D convolution filters.
The following studies have directed their attention toward the adaptation of sampling grid positions of convolution filters~\cite{tateno2018distortion}.
SphereNet~\cite{coors2018spherenet}, for instance, is a notable work that has demonstrated the effectiveness of this methodology in tasks such as classification and object detection.
Zhao \textit{et al.}~\cite{zhao2018distortion} further leveraged a non-regular grid for each pixel based on its distortion level and convolved the sampled grid using square kernels shared by all pixels, facilitating end-to-end training.
Apart from adapting 2D CNN kernels, another approach involves establishing spherical convolution kernels to achieve rotational equivariance and invariance.
For instance, Esteves \textit{et al.}~\cite{esteves2018learning} made the first attempt to implement spherical harmonic domain CNN filters via group convolutions.
Meanwhile, Yang \textit{et al.}~\cite{yang2020rotation} and Perraudin \textit{et al.}~\cite{perraudin2019deepsphere} took a different approach by leveraging a graph to represent the spherical image and achieving resilient isometric equivariance transformation through via hierarchical equal area isolatitude pixelization.

\subsection{Omni-Directional Image Super-Resolution}
Preliminary ODR-SR models adopted a naive strategy that straightforwardly adopted the checkpoints of planar image SR models and fine-tuned them on the ODIs, resulting in limited performance~\cite{ozcinar2019super}.
LAU-Net~\cite{deng2021lau} proposed the first ODI-SR approach that considered the non-uniform sampling characteristics of ODR images. Specifically, a tile-based strategy was introduced, where the ERP images were first split into several regions according to latitude, and the upsampling factors were adaptively assigned.
Yoon \textit{et al.}~\cite{yoon2022spheresr} explored the usage of icosahedron projection to minimize sphere-to-planar distortion and proposed a new kernel weight-sharing scheme aligned with the icosahedron projection.
Although these methods successfully enhance the performance of ODI-SR by incorporating the spherical characteristics of ODI, the specially designed models involve tedious operations, posing significant obstacles to both running speed and practical implementation and hindering further research of ODI-SR.
Moreover, other works have tended to incorporate sphere-to-planar knowledge as side information into the feature extraction process, such as the stretching ratio or distortion map~\cite{yu2023osrt}.
However, the naive incorporation method and the sole focus on the feature extraction process prevent these approaches from achieving state-of-the-art performance.

\subsection{Implicit Image/Neural Representation}
The implicit neural representation (INR) originates from the modeling of 3D object-shape surfaces~\cite{atzmon2020sal,chen2019learning,michalkiewicz2019implicit} and involves leveraging a multi-layer perceptron function to map coordinates to the signal.
Inspired by INR's favorable capacity for recovering fine details of shapes, it has started to be employed in representing planar images, such as in image restoration~\cite{anokhin2021image,dupont2022generative,skorokhodov2021adversarial} and image compression~\cite{strumpler2022implicit}.
Preliminary INR involves learning a specific mapping function for a single image, raising concerns about inference time in practical implementation scenarios.
As such, recent works tend to explore an implicit representation function space shared by different images~\cite{chen2021learning,song2023ope,nguyen2023single}, where the extracted deep features are also regarded as inputs alongside the coordinates for mapping to the signals, denoted as implicit image function.
In the context of image SR, the implicit image function has obtained significant progress, especially in arbitrary-scale image SR, such as SIREN~\cite{sitzmann2020implicit} and LIIF~\cite{chen2021learning}, since the coordinate-to-signal function is well-aligned with continuous image representation.
Furthermore, in the ODR-SR domain, SphereSR~\cite{yoon2022spheresr} made the first effort to incorporate INR into spherical coordinates to achieve arbitrary-scale SR processing.
However, the spherical implicit image function is performed on a reprojected icosahedron surface and involves a complex convolutional kernel modification process to align with the icosahedron ODI pixel representation. This inevitably limits the performance and inference speed.

\section{Method}
\subsection{Motivation and Overview}
As described in the background information, existing studies on ODI-SR are still insufficient. 
Shortages of exiting ODI-SR routes are summarized as follows, 
1) Tile-based methods disrupt the spatial continuity of the input image, failing to comprehensively understand the scene context and provide satisfactory performance.
2) Reprojection-based methods involve specially designed convolution kernels to align with the corresponding pixel representation approach, resulting in a cumbersome and time-consuming inference process.
3) Sphere-to-planar prior-guided methods successfully address the above issues. However, the naive incorporation of prior information and the lack of consideration for the image reconstruction process prevent these methods from achieving state-of-the-art performance.

Motivated by this, this paper proposes a novel Fast and Arbitrary-Scale ODI Super-Resolution (FAOR) method that adapts the implicit image function to the ERP domain, incorporating spherical geometric priors into both the latent representation and image reconstruction processes.
Bearing both fast inference speed and high SR performance in mind, the spherical geometric priors are integrated via specially designed lightweight modules.
Moreover, although numerous projection methods exist for ODI representation, such as cube map, fisheye, and polyhedron, it has to be mentioned that almost all raw ODIs are recorded and stored in ERP format.
With other projection types derived from ERP, patterns across them are reusable when distortions are correctly rectified.
%
%
The architecture of the proposed FAOR framework is illustrated in Fig.~\ref{fig:framework}, with its main modules highlighted as follows:
\begin{itemize}
    \item \textbf{Sphere-to-planar-Aware Feature Encoder (SAFE).} The SAFE is responsible for extracting the latent representation  $Z\in \mathbb{R}^{H\times W\times D}$ of the input LR ODI $X\in \mathbb{R}
^{H\times W\times 3}$, where $D$ denotes the feature map channels. 
    In general, the SAFE is characterized by an Affine-Transformation-based Feature Modulating (ATFM) module and a Cross-Attention (CA)-based feature enhancement module. These are specially designed to incorporate spherical geometric priors and further enhance sphere-to-planar awareness, respectively.
    \item \textbf{Affine-Transformation-based Feature Modulating (ATFM).} The ATFM module jointly leverages a pixel-wise stretching ratio map  $M_{d}$ and an instance segmentation map $M_{s}$ to generate a set of affine parameters, thereby obtaining sphere-to-planar aware latent representations.
    \item \textbf{Spherical Geodesic-based Implicit Neural Function (SGIF).} 
    To align the image representation process with the spherical characteristics, the SGIF leverages a spherical geodesic-based resampling method, obtaining the latent representation $\hat{Z}\in \mathbb{R}^{sH\times sW\times D}$ of the HR-ODI, where $s$ is the arbitrary upsampling scale.
    The $\hat{Z}$ and the corresponding spherical coordinates $[\hat{\boldsymbol{\theta}},\hat{\boldsymbol{\phi}} ]$ are subsequently fed to the continuous implicit image function, obtaining the HR-ODI $Y\in \mathbb{R}^{sH\times sW\times 3}$.
\end{itemize}
Details of the modules will be presented in the following subsections.
\subsection{Sphere-to-Planar-Aware Feature Encoder}
The proposed SAFE is characterized by $L$ duplicated feature encoder blocks, with details illustrated in Fig.~\ref{fig:framework}. 
To gain intuition into the encoding process, denote the input feature of the $l-th$ encoder block as $f_{l}$, it would be first fed to a Layer Norm (LN) layer and subsequently to the ATFM module to generate the sphere-to-planar aware features $f_{at}$, 
\begin{equation}
    f_{ln}=\text{LN}(f_{l}),\,\,f_{at}=\text{ATFM}(f_{ln}),
\end{equation}
where the ATFM applies the affine transforms on the input features according to the spherical geometric and semantic priors, corresponding details will be presented in the following subsection.
To further enhance awareness of the sphere-to-planar distortion, we incorporated a Cross-Attention (CA) module that conducts channel-wise attention between features with and without the affine transform.
In special, consider the $q_{ln}$ as the query extracted from $f_{ln}$,  $k_{at}$ and $v_{at}$ are the key and value associated with $f_{at}$, the CA process can be formulates as,
\begin{equation}
f_{ca}=\text{MLP}(\frac{q_{ln}\times  k_{at}}{\sqrt{d_k}}v_{at}),
\end{equation}
where $d_k$ is a scaling factor, MLP denotes the multi-layer perceptron layer.
Moreover, a skip connection bypasses the ATFM and CA modules, and an additional residual block containing an LN layer and an MLP layer is incorporated to further boost training performance.
\begin{equation}
\begin{split}
   &\widetilde{f_{ca}}= f_{l}\oplus f_{ca},\\
   &f_{l+1}= \widetilde{f_{ca}}\oplus \text{MLP}(\text{LN}(\widetilde{f_{ca}})),
\end{split}
\end{equation}
where $f_{l+1}$ denotes the output of the $l-th$ encoder block.
\subsection{Affine-Transformation-based Feature Modulating}
The distortion pattern associated with sphere-to-planar projections in ERP is characterized by stronger stretching ratios at higher latitudes, whereas the stretching ratio along the longitude remains constant, and can be roughly formulated as,
\begin{equation}
D_{ERP} (h, w)= D_{ERP} (\theta,\phi)=\text{cos}(\phi);
\end{equation}
where $(h, w)$ and $(\theta,\phi)$ are equivalent coordinates in planar and spherical coordinate systems, respectively.
To represent the nonuniform stretching artifact, this paper explores the use of affine transformations, which are capable of performing complex deformations, such as rotation and zooming, with streamlined computations.
\begin{figure}[t]
    \hfill
    \centering
    \includegraphics[width=0.42\textwidth]{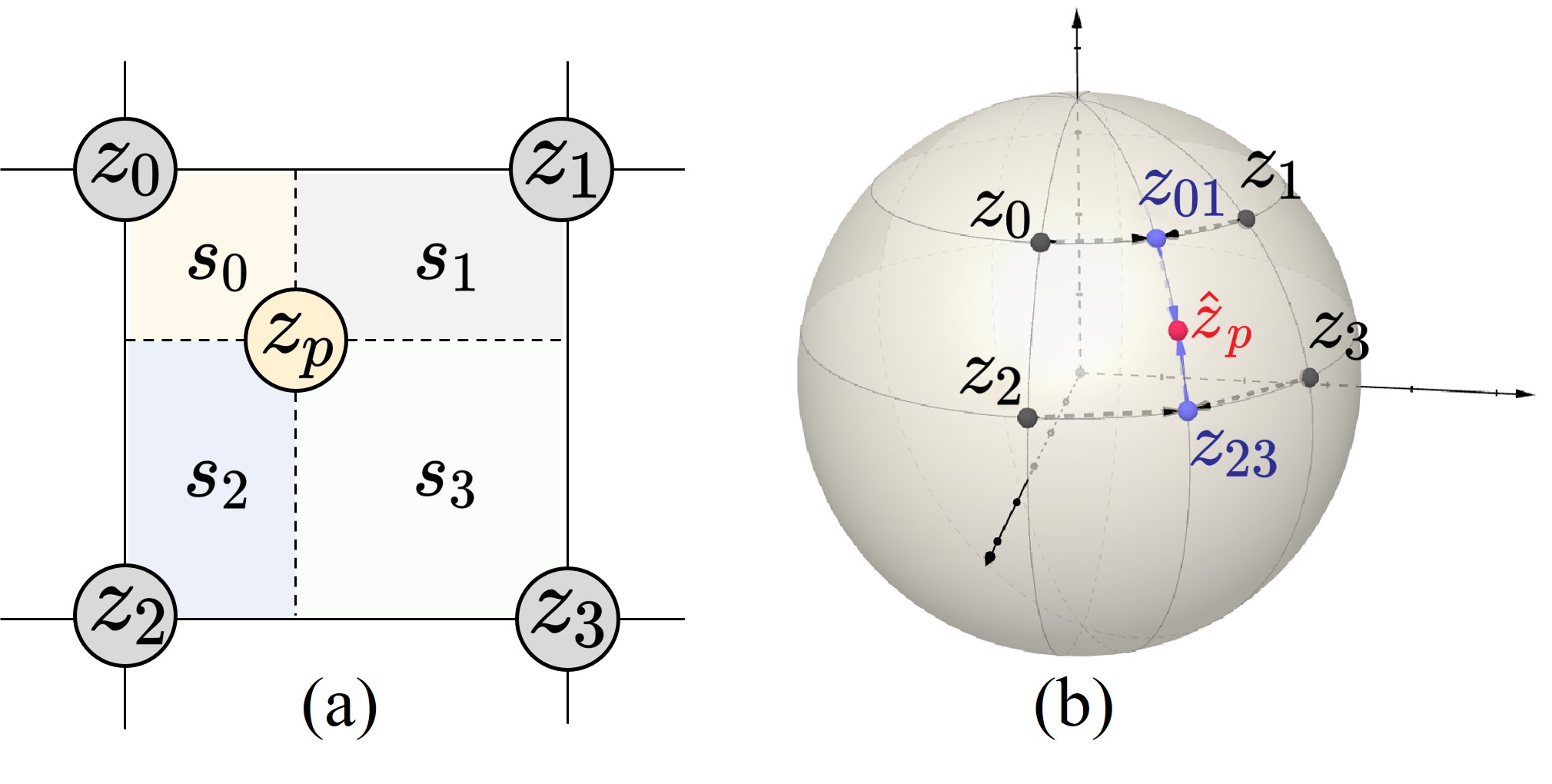}
    \caption{(a) The normalized interpolation method in LIIF, leveraging the areas of $s_0$, $s_1$, $s_2$, and $s_3$ to obtain normalized weights for $z_p$ interpolation. (b) The proposed spherical geodesic-based interpolation.}
    \label{fig:resampling_compare}
\end{figure}

In general, the ATFM module takes additional offline-generated priors as input to generate a set of element-wise affine parameters $(\alpha,\beta)$, performing the affine transformations on the input features. 
Details of the proposed module are illustrated in Fig.~\ref{fig:framework} (c), where $f_{ln}$ and $f_{at}$ denote the input and output features of the module, respectively.

As for the incorporated priors, a straightforward stretching ratio map  $M_{d}$ that reflects the pixel-wise sphere-to-planar distortion is generated by, 
\begin{equation}
M_{d} (h, w)=255\times \text{cos}(\frac{h+0.5-H/2}{H}\pi).
\end{equation}
Moreover, we adopt an instance segmentation map 
$M_{s}$ from~\cite{zhang2022bending} to enhance awareness of spherical characteristics from the semantic deformation perspective.
By constraining the perception perspective to the instance level, 
$M_{s}$ is expected to provide a more thorough and accurate understanding of the sphere-to-planar distortion for the ATFM model.
\begin{figure*}[ht]
    \centering
    \includegraphics[width=0.9\textwidth]{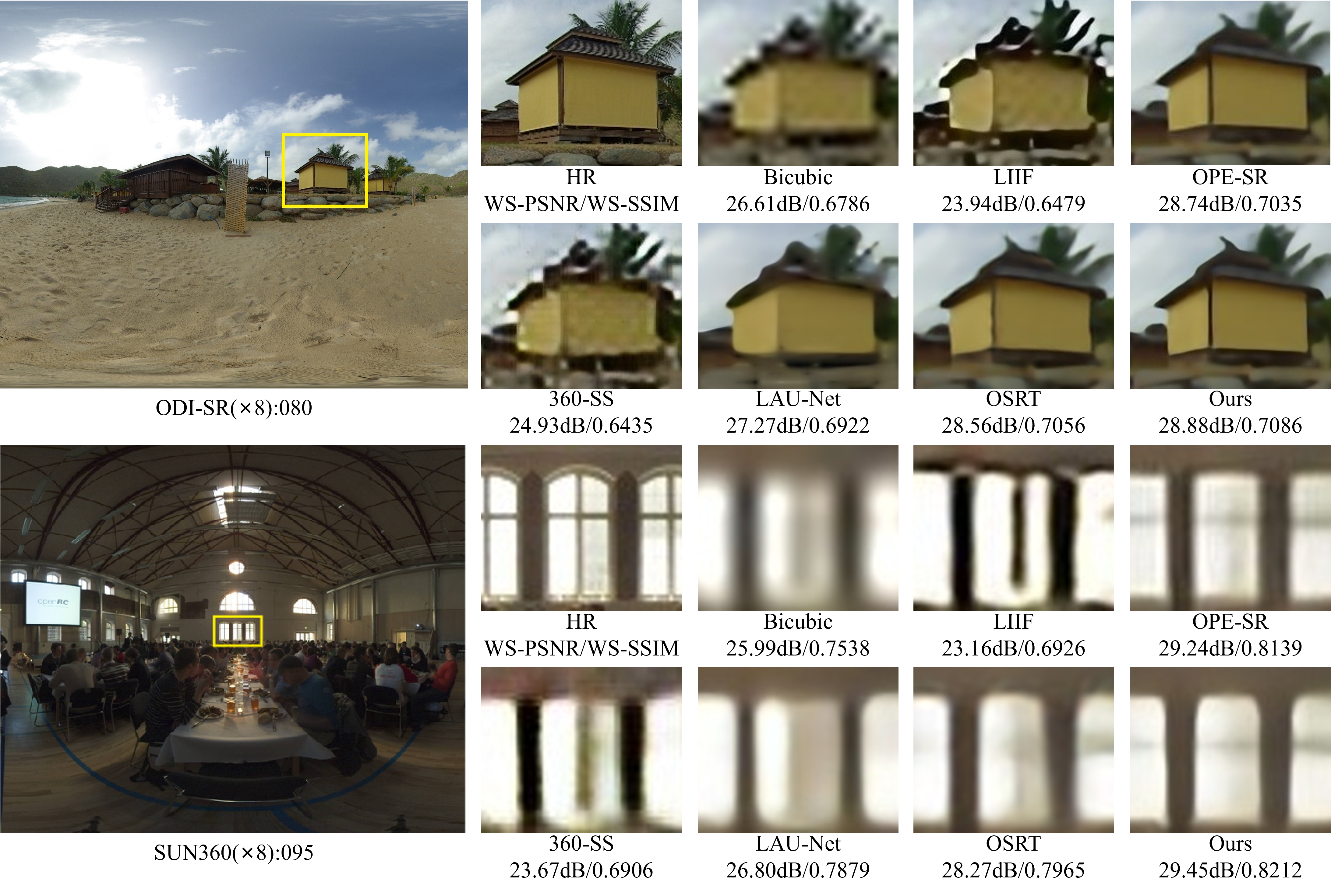} 
    \caption{Visual comparisons of $\times 8$ SR results on ODI-SR and SUN360 testing sets.}
    \label{fig:sr-odi2}
\end{figure*}
%
%
Thus, the entire ATFM process can be formulated as,
\begin{equation}
  f_{at}=\text{ATFM}(f_{ln},M_{s},M_{d})=\alpha \odot f_{ln}+\beta, 
\end{equation}
where $f_{ln}$ and $f_{at}$ denote the input and output features of the ATFM module, respectively.
\begin{table*}[t]
\centering
\begin{tabular}{>{\centering\arraybackslash}p{2cm}|cc|cc|cc|cc}
\toprule
Dataset & \multicolumn{8}{c}{ODI-SR} \\
\midrule
\multirow{2}{*}{Method} & \multicolumn{2}{c|}{$\times 2$} & \multicolumn{2}{c|}{$\times 4$} & \multicolumn{2}{c|}{$\times 8$} & \multicolumn{2}{c}{$\times 16$} \\
 & WS-PSNR & WS-SSIM & WS-PSNR & WS-SSIM & WS-PSNR & WS-SSIM & WS-PSNR & WS-SSIM \\
\midrule
Cubic & 27.61 & 0.8156 & 24.95 & 0.6923 & 19.64 & 0.5908 & 17.12 & 0.4332 \\
LIIF & 27.34 & 0.8214 & 22.29 & 0.6626 & 19.45 & 0.5692 & 17.59 & 0.5231 \\
OPE-SR & 29.20 & 0.8522 & \underline{26.48} & 0.7435 & \underline{24.50} & 0.6543 & \textbf{22.82} & 0.5992 \\
360-SS & 27.14 & 0.8095 & 23.20 & 0.6613 & 21.65 & 0.6417 & 19.65 & 0.5431 \\
LAU-Net & 29.33 & 0.8633 & 26.34 & 0.7352 & 24.36 & \textbf{0.6602} & 22.07 & 0.5901 \\
OSRT & \underline{29.61} & \underline{0.8700} & 26.26 & \underline{0.7443} & 24.24 & 0.6532 & 22.49 & \textbf{0.6030} \\
Ours & \textbf{30.12} & \textbf{0.8796} & \textbf{26.69} & \textbf{0.7560} & \textbf{24.58} & \underline{0.6581} & \underline{22.78} & \underline{0.6008} \\
\midrule
Dataset & \multicolumn{8}{c}{SUN360 Panorama} \\
\midrule
Cubic & 28.01 & 0.8321 & 24.90 & 0.7083 & 19.72 & 0.5403 & 17.56 & 0.4638 \\
LIIF & 28.20 & 0.8424 & 22.17 & 0.6708 & 19.07 & 0.5706 & 16.59 & 0.5258 \\
OPE-SR & 30.67 & 0.8782 & \underline{27.42} & \underline{0.7809} & \underline{24.38} & \underline{0.6848} & \underline{22.50} & \underline{0.6161} \\
360-SS & 27.97 & 0.8294 & 23.19 & 0.6725 & 21.48 & 0.6352 & 19.62 & 0.5308 \\
LAU-Net & 29.11 & 0.8555 & 26.65 & 0.7479 & 24.02 & 0.6708 & 21.82 & 0.5824 \\
OSRT & \underline{31.20} & \underline{0.8958} & 26.93 & 0.7780 & 24.24 & 0.6533 & 22.31 & \textbf{0.6211} \\
Ours & \textbf{32.06} & \textbf{0.9096} & \textbf{27.66} & \textbf{0.7950} & \textbf{24.77} & \textbf{0.6914} & \textbf{22.53} & \textbf{0.6211} \\
\bottomrule
\end{tabular}
\caption{Quantitative comparison with state-of-the-art SR methods on ODI-SR and SUN360 testing sets. The bold and underlined font indicate the best and second-best results, respectively.}
\label{table:result_odisr}
\end{table*}
\subsection{Spherical Geodesic-based Implicit Image Function}
In general, the implicit image function involves establishing a mapping function from latent vectors and corresponding coordinates to pixel values.
With well-established mapping functions, the latent vectors closest to the targeted HR coordinate are fed into the mapping function and subsequently passed through a normalized interpolation, obtaining the targeted HR pixel values.

To align the image reconstruction process with spherical characteristics, the SGIF specifically incorporates a geodesic-based spherical resampling function. 
An intuitive comparison between the existing planar resampling function and the proposed spherical-oriented approach is illustrated in Fig.~\ref{fig:resampling_compare}.
Moreover, we propose a \textit{resampling-then-representation} strategy that directly performs geodesic-based spherical resampling on the latent rather than on the pixel representation.
This approach enables the implicit image function to be executed only once per HR pixel prediction, for the sake of fast inference speed.

In particular, with the obtained latent representation of the LR-ODI $Z$, we first obtain $H\times W$ latent vectors $[z_{11},z_{12},...,z_{HW}]$, 
where $z_{hw}\in \mathbb{R}^{1\times 1\times D}$ corresponding to the spherical coordinate $[\theta_{h},\phi_{w} ]$.
Subsequently, the HR coordinates $\widehat{\boldsymbol{\theta}}=[\widehat{\theta}_{1}, \widehat{\theta}_{2},...,\widehat{\theta}_{sH}]$, and $\widehat{\boldsymbol{\phi}}=[\widehat{\phi}_{1}, \widehat{\phi}_{2},...,\widehat{\phi}_{sW}]$.
are calculated and contributed to obtain the corresponding latent vectors $\widehat{Z}=[\widehat{z}_{11}, \widehat{z}_{12},..., \widehat{z}_{sHsW}]$ via spherical linear interpolation of $Z$.  
%
%
For simplicity, denote the target spherical coordinate as $[\widehat{\theta}_{p},\widehat{\phi}_{p}]$, the four nearby reference latent vectors as $z_{0}$, $z_{1}$, $z_{2}$, $z_{3}$, as illustrated in Fig.~\ref{fig:resampling_compare}(b).
The spherical linear interpolation is conducted to obtain the HR latent vector $\widehat{z}_p$ corresponding to the target coordinate.

Specifically, we introduce a geodesic-based spherical linear interpolation since the ERP pixels within each row (colum) share the same latitude (longitude)~\cite{fatelo2021mobility}.
We resample $z_{0}$ and $z_{1}$ to $z_{01}$, $z_{2}$ and $z_{3}$ to $z_{23}$,
\begin{equation}
\begin{split}
z_{01}= \frac{\text{sin}(1-t_{01})\delta_{01}}{\text{sin} \delta_{01}}z_{0}+\frac{\text{sin} t_{01} \delta_{01}}{\text{sin} \delta_{01}}z_{1},\\
z_{23}= \frac{\text{sin}(1-t_{23})\delta_{23}}{\text{sin} \delta_{23}}z_{2}+\frac{\text{sin} t_{23} \delta_{23}}{\text{sin} \delta_{23}}z_{3},
\end{split}
\end{equation}
where $\delta_{01}$ denote the longitude angle  subtended by $z_{0}$ and $z_{1}$, as well as $\delta_{23}$.
Since the $z_{01}$ and $z_{23}$ are with the same longitude, the latent vector at the targeted coordinate can be obtained by,
\begin{equation}
\widehat{z}_{p}= \frac{\text{sin}(1-t_{02})\delta_{02}}{\text{sin} \delta_{02}}z_{01}+\frac{\text{sin} t_{02} \delta_{02}}{\text{sin} \delta_{02}}z_{23},
\end{equation}
where the $\delta_{02}$ and $t_{02}$ are calculated based on the latitude of $z_0$ and $z_{2}$.

As illustrated in Fig.~\ref{fig:framework}, 
the obtained latent vector $\widehat{z}_{p}$ and its spherical coordinate would be fed to the SGIF, predicting the corresponding pixel value $y_p$,
\begin{equation}
y_p= \text{SGIF}(\widehat{z}_{p},[\widehat{\theta}_{p},\widehat{\phi}_{p}]).
\end{equation}
Along this vein, the predicted HR ODI image $Y=[y_{11},y_{12},...,y_{sHsW}]$ can be obtained in a pixel-by-pixel manner.
%
%
\subsection{Training and Supervision}
A specifically designed self-supervised training strategy is employed, allowing the proposed SR model to be trained just once to gain the capacity for arbitrary-scale ODI-SR.
In detail, the pristine HR ERP images are first cropped into a series of patches with resolution $128r \times 128 r$ as ground truths, where $r$ denotes random floating-point numbers.
Subsequently, these ground-truth patches are downsampled to a uniform size of $128 \times 128$ using cubic interpolation, creating LR-HR pairs with randomized scales.
Moreover, to achieve arbitrary-scale SR capacity within a single training approach, a fixed number of $128 \times 128$ pixels are randomly selected from different resolution ground-truth patches and contributed to the SGIF optimization.
By constraining the selected pixel coordinates and $r$ to uniform distributions, the SGIF is supplied with sufficient latent vector-pixel pairs at arbitrary coordinates.
Moreover, the entire model optimization process is supervised by pixel-wise $\mathcal{L}_{1}$ loss.

\section{Experiment}
In this section, extensive experiments are conducted to demonstrate the superiority of the proposed method against the state-of-the-art regarding inference speed and perceptual quality. 
Furthermore, comprehensive ablation studies have demonstrated the effectiveness of our design.
\subsection{Experimental Setting}
\textit{Dataset:} In the experimental stage, the ODI-SR~\cite{deng2021lau} and SUN360~\cite{xiao2012recognizing} datasets are employed.
In particular, the training sets of ODI-SR and SUN360 are jointly leveraged for training, comprising a total of 1151 ODIs with a resolution of $2048 \times 1024$.
The performance on their testing sets, each containing 100 images, is separately reported.
During the training process, corresponding LR versions are down-sampled via Cubic.

\textit{Benchmark:} As for the benchmark, six SR models with various architectures and mechanisms are employed to provide a comprehensive comparison. This includes three state-of-the-art ODI-SR models: 360-SS~\cite{ozcinar2019super}, LAU-Net~\cite{deng2021lau}, and OSRT~\cite{yu2023osrt}; two cutting-edge 2D SR models, LIIF~\cite{chen2021learning} and OPE-SR~\cite{song2023ope}; and the conventional cubic interpolation method.
Specifically, the 360-SS, LAU-Net, and OSRT models perform fixed-scale SR and are retrained for specific scales. Meanwhile, the OPE-SR and LIIF models achieve arbitrary-scale SR.
All the learning-based models are retrained with the same dataset as ours to ensure fair comparisons.
Regarding the evaluation criteria, WS-PSNR~\cite{sun2017weighted} and WS-SSIM~\cite{zhou2018weighted} have been adopted.

\textit{Implementation Details:} We employ the Adam optimizer~\cite{kingma2014adam} with an initial learning rate of $10^{-4}$, which will be halved at the 30k, 50k, 100k, and 400k-th iterations. 
The $L$ indicating the number of SAFEB is set to 36.
The networks are implemented in PyTorch 1.10.2 and Python 3.8.16. Training is performed on a machine equipped with 8 NVIDIA RTX 3090 GPUs.

\subsection{Experimental Results}
\textit{Quantitative Results:} Comparisons between the proposed FAOR and the employed anchors are made at $\times2$, $\times4$, $\times8$, and $\times16$ scale SR tasks, corresponding results regarding reconstruction performances on testing set of ODI-SR and SUN360 are provided in Table~\ref{table:result_odisr}.
In particular, ODI-oriented SR methods have demonstrated overwhelming advantages over 2D SR models, underscoring the necessity of incorporating ODI characteristics into SR models.
Moreover, compared to the state-of-the-art ODI-SR method, OSRT, our proposed method achieves an average gain of 0.39 dB in WS-PSNR and 0.006 in WS-SSIM on the ODI-SR dataset, and a 0.59 dB gain in WS-PSNR and 0.017 in WS-SSIM on the SUN 360 dataset.

\textit{Qualitative Comparison:} A set of comparisons on $\times 8$ scale SR is provided in Fig.~\ref{fig:sr-odi2}.
As shown, the overwhelming advantages of the proposed FAOR over LIIF, 360-SS, and LAU-Net can be easily observed, as FAOR is capable of providing more natural object structuring and greatly improving the recovery of fine textures, \textit{e.g.,} the house and tree regions in the above figure.
Meanwhile, LAU-Net and 360-SS failed to recover detailed information and suffered from significant erasing artifacts.
%
%
%
%
Moreover, in complex scenarios, the limitations of cutting-edge methods such as OPE-SR and OSRT become evident, with failures to reconstruct the texture details in the window regions, inevitably resulting in blurring artifacts.
Meanwhile, the proposed FAOR continues to recover fine texture details effectively, owing to the incorporation of semantic and pixel-wise sphere-to-planar-aware priors.

\textit{Inference Speed Comparison:} A comprehensive comparison regarding inference running speed, model parameters, and reconstruction performance on $\times 8$ scales is provided in Fig.~\ref{fig:Arbitrary-Performance} (b).
Specifically, 360-SS achieves a faster running speed as it leverages a naive adaptation method that directly adapts 2D image SR models to the ODI domain using GAN-based supervision.
However, the lack of consideration for spherical characteristics also results in poor visual representation.
Compared to LIIF, our method achieves a faster running speed despite having more parameters, primarily due to the incorporation of the \textit{resampling-then-representation} strategy.
Unlike the vanilla local implicit image representation in LIIF, which requires executing the implicit function four times to predict one pixel, the \textit{resampling-then-representation} strategy further boosts the running speed while maintaining the reconstruction performance.
Moreover, compared to the cutting-edge OSRT, the proposed FAOR has a similar model size but operates much faster.
The underlying reason lies in this paper's exploration of a new path for spherical characteristics incorporation, utilizing the streamlined affine transformation for sphere-to-planar awareness instead of complex spherical operators and reprojection processing, thereby striking a good balance between inference speed and SR performance.
\subsection{Ablation Study}
Extensive ablation studies are conducted to demonstrate the effectiveness of the incorporated spherical geometric priors, in both the SAFE and SGIF.
The corresponding results and analysis are presented in this subsection.
\subsubsection{Ablation of Incorporated Priors in SAFE}
In the latent representation stage, the pixel-wise stretching ratio map $M_{d}$ and the instance segmentation map $M_{s}$ are adopted to provide a comprehensive understanding of the sphere-to-planar distortion.
Herein, we examine their contributions to SR performance on the ODI-SR dataset by gradually ablating them, while keeping all other implementation details strictly the same as in our full version.
The corresponding results are listed in Table~\ref{table:abl_msmd}.
The effectiveness of the $M_{d}$ and $M_{s}$  is evident, as they both contribute to an overall performance gain across multiple SR scales.

\begin{table}[t]
\centering
\setlength{\tabcolsep}{0.8mm}{
\begin{tabular}{>{\centering\arraybackslash}p{2cm}|ccccc}
\toprule
Method & Metrics & $\times 2$ & $\times 4$ & $\times 8$ & $\times 16$ \\
\midrule
\multirow{2}{*}{Ours} & WS-PSNR & \textbf{30.12} & \textbf{26.69} & \textbf{24.58} & \textbf{22.78} \\
  & WS-SSIM & 0.8796 & \textbf{0.7560} & \textbf{0.6581} & \textbf{0.6008} \\
\midrule
\multirow{2}{*}{$w/o$ $M_{d}$} & WS-PSNR & 30.11 & 26.67 & 24.47 & 22.69 \\
  & WS-SSIM & \textbf{0.8850} & 0.7558 & 0.6575 & 0.6004 \\
\midrule
\multirow{2}{*}{$w/o$ $M_{d}+M_{s}$} & WS-PSNR & 29.79 & 26.45 & 24.35 & 22.64 \\
  & WS-SSIM & 0.8741 & 0.7482 & 0.6502 & 0.5949 \\
\bottomrule
\end{tabular}
}
\caption{Ablation study of incorporated priors in SAFE. All models are tested on ODI-SR testing set. }
\label{table:abl_msmd}
\end{table}
\subsubsection{Ablation of Spherical Geodesic-based Resampling Function}
To demonstrate the effectiveness of the incorporated geodesic-based resampling function, we replace it with the normalized interpolation method in LIIF (denoted as `\textit{w/o sphere}'), corresponding comparison results on the ODI-SR dataset are provided in Table~\ref{table:abl_sphereinter}. 
The benefit of the incorporated spherical geometric prior is easily observed, resulting in an improvement of 0.12 dB in WS-PSNR and 0.004 in WS-SSIM, respectively.
\begin{table}[t]
\centering
\setlength{\tabcolsep}{0.8mm}{
\begin{tabular}{>{\centering\arraybackslash}p{2cm}|ccccc}
\toprule
Method & Metrics & $\times 2$ & $\times 4$ & $\times 8$ & $\times 16$ \\
\midrule
\multirow{2}{*}{Ours} & WS-PSNR & \textbf{30.12} & \textbf{26.69} & \textbf{24.58} & \textbf{22.78} \\
  & WS-SSIM & \textbf{0.8796} & \textbf{0.7560} & \textbf{0.6581} & \textbf{0.6008} \\
\midrule
\multirow{2}{*}{$w/o$ $sphere$} & WS-PSNR & 29.95 & 26.56 & 24.42 & 22.78 \\
  & WS-SSIM & 0.8761 & 0.7510 & 0.6533 & 0.5974 \\
\bottomrule
\end{tabular}
}
\caption{Ablation study of spherical geodesic-based resampling function. }
\label{table:abl_sphereinter}
\end{table}

\subsection{Conclusion}
Aiming to perform fast and arbitrary-scale omnidirectional image super-resolution, this paper adapts the implicit image function from the planar image domain to the ERP image domain by incorporating multiple spherical geometric priors into both the latent representation and image reconstruction processes. 
Specifically, to obtain sphere-to-planar aware latent representations, we leverage a pixel-wise stretching ratio map and an instance segmentation map to enhance sphere-to-planar awareness from both pixel-wise and semantic perspectives.
Additionally, we introduce a spherical geodesic-based resampling function to align the implicit image function with spherical characteristics. All incorporated modules are specially designed with respect to both complexity and efficiency. As a result, the proposed method achieves superior performance compared to state-of-the-art methods, with significantly faster inference speeds.

%

\section*{Acknowledgments}
This work was supported in part by the National Natural Science Foundation of China under Grants 62371310, in part by the Guangdong Basic and Applied Basic Research Foundation under Grant 2023A1515011236 and 2024A1515010454, in part by the Basic and Frontier Research Project of PCL, in part by the Major Key Project of PCL, and in part by by the Open Research Fund from Guangdong Laboratory of Artificial Intelligence and Digital Economy (SZ)  under Grant No. GML-KF-24-27.

\bibliography{aaai25}

\end{document}